# CONTEXT-AWARE DEEP LEARNING FOR MULTI-MODAL DEPRESSION DETECTION

*Genevieve Lam[1], Huang Dongyan[2,3], and Weisi Lin[1]*

[1]School of Computer Science and Engineering, Nanyang Technological University, Singapore
[2]Institute of Infocomm Research, A*STAR, Singapore
[3]UBTech, Shenzhen City, 518055, P. R. China
{C150110@e.ntu.edu.sg, WSLin@ntu.edu.sg},{dongyan.huang@ubtrobot.com}

**ABSTRACT**

In this study, we focus on automated approaches to detect depression from clinical interviews using multi-modal machine learning (ML). Our approach differentiates from other successful ML methods such as context-aware analysis through feature engineering and end-to-end deep neural networks for depression detection utilizing the Distress Analysis Interview Corpus. We propose a novel method that incorporates: (1) pre-trained Transformer combined with data augmentation based on topic modelling for textual data; and (2) deep 1D convolutional neural network (CNN) for acoustic feature modeling. The simulation results demonstrate the effectiveness of the proposed method for training multi-modal deep learning models. Our deep 1D CNN and Transformer models achieved state-of-the-art performance for audio and text modalities respectively. Combining them in a multi-modal framework also outperforms state-of-the-art for the combined setting.

## 1. INTRODUCTION

Major depressive disorder (MDD) is a highly prevalent condition affecting around 350 million across the globe. Approximately 76% of sufferers seek help and diagnosis for their depression through a general practitioner (GP) [1]. GPs use the Patient Health Questionnaire (PHQ), a standard method for diagnosis of depression [2]. The questionnaire asks individual questions on whether they feel tired, struggle with sleeping, have a poor appetite, enjoy being with family, etc. to help the GP obtain a self-reported depression criterion.

However, a GP's judgments often result in some false negatives and/or false positives even with the PHQ screening test [3]. A passive automated monitoring of human communication could solve these constraints and provide a better alternative to screening for depression.

Successful machine learning approaches to depression detection have relied on feature engineering such as topic modelling based on the subject's responses to specific types of questions [4], or on purely data-driven end-to-end trained deep neural network models [5, 6] which attempt to exploit global and/or time-varying statistics.

In this work, we aim to evaluate multi-modal approaches to automatic depression detection using the DAIC WOZ corpus [7], comparing context-aware approaches to fully data-driven approaches based on deep learning. Our ultimate goal is to propose new techniques for effectively augmenting a deep neural network's ability to model non-linear relationships between multi-modal signals with contextual prior knowledge. Based on the success of topic modelling to obtain discriminative features for simple models such as decision trees [8], we train deep neural networks using participants' responses to specific questions which were found to be highly predictive of depression [4].

A key challenge to training such data-driven systems is the limited availability and high-class imbalance of data. Most clinical corpora for depression have far more samples from non-depressed individuals compared to depressed ones. As a result, past work has focused on resolving class imbalance by data sampling methods [5, 9]. We propose an alternative solution by using a novel data augmentation framework based on topic modelling. This helps us create a larger, more balanced dataset and increases the robustness of trained models.

We obtain strong empirical results for audio, text and multi-modal settings when training deep learning models using the proposed data augmentation framework, pointing to the promise of combining context-aware and data-driven methods for the automatic detection of depression.

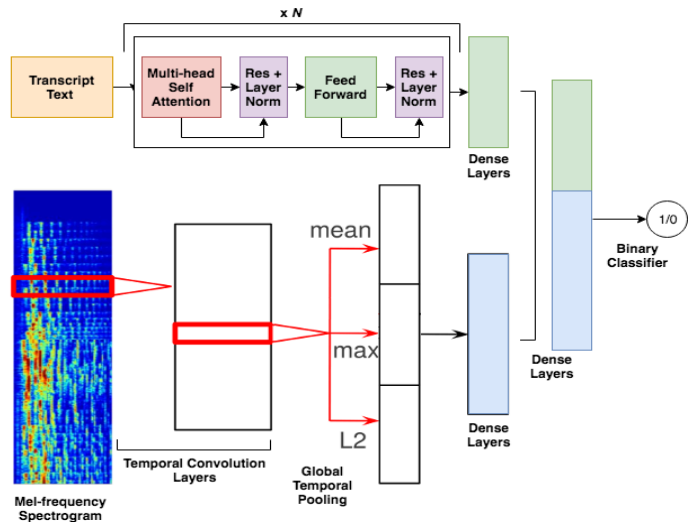

Figure 1: **The proposed multi-modal model** consisting of a Transformer (top) and CNN (bottom) models. The features from both models are concatenated and passed to a feedforward model that predicts whether an individual is depressed or not.

## 2. RELATED WORK

### 2.1. Feature Engineering based approaches (context-aware)

For the DAIC-WOZ dataset, past work has revolved around extracting features from a patient's answers against certain types of topics and questions which were more predictive of depression [4, 8]. Yang et *al.* obtained good performance using manually chosen questions and modeled features from their answers in a decision tree [8]. Other approaches used a multi-modal ensemble model based on the question type asked [4, 9].

The work in [9] is closest to ours. They used deep neural networks for multiple modalities with model re-sampling to tackle data imbalance. Our framework differs from their approach due to data augmentation based on topic modelling as well as our usage of more powerful deep learning models with transfer learning.

### 2.2. Data-driven models (context-independent)

Several deep learning approaches have obtained good empirical performance using only the semantic and/or prosodic features obtained from participants' replies without modelling it to the specific topics.

One-dimensional Convolutional Neural Networks (1D CNNs) for obtaining features from audio signals with an Long-Short Term Memory (LSTM) for temporal fusion were proposed in *DepAudioNet* [5]. We utilize similar 1D CNNs in our audio-based models but differ from [5] in our use of global pooling layers instead of more computationally expensive LSTMs.

In light of the limited availability of training data, transfer learning has been used to improve detection depression in one social media (Weibo) using cross-domain deep neural networks and large-scale data from another social media (Twitter) [10]. Taking a different approach to transfer learning for text data from interview transcripts, we obtain strong empirical results by fine-tuning the recently proposed Transformer model [11] which was pre-trained on the unsupervised task of language modelling.

Finally, bidirectional LSTM models for audio and text modalities were used in [6] to obtain state-of-the-art performance by purely data-driven models.

## 3. FEATURES

The audio and text of 189 participants from the DAIC-WOZ dataset was utilized for this study [7]. In this dataset, participants undergo clinical interviews conducted by an avatar named Ellie through a human interviewer in a "Wizard of Oz" approach. Text and audio data capture the interaction between Ellie and a participant. Each interview ranges between 7-33 minutes. The dataset was split into 3 sets - training (57%, 107 participants), development (19%, 35 participants) and test (25%, 47 participants) sets as specified by [7]. However, the test set did not come with annotations. Hence, testing is done on the development set instead, following [6].

For obtaining input data for our models, we utilized the raw text transcripts provided in the dataset and converted audio data (only participant segments) to Mel-frequency spectrograms.

Mel-frequency was used as shown in [5, 12], as it provides audio features with low-level audio descriptors. All Mel-spectrograms were log-scaled to minimize the mean squared error (MSE) of the predictions.

## 4. MODELS AND APPROACHES

### 4.1. Naïve Topic Modelling

Based on extensive manual topic classification in [4], we further categorized the topics into 7 main topics most related to the PHQ8 questions: if they have interests, if they have been having a good sleep, do they feel depressed, do they feel like a failure, what is their personality, have they been diagnosed with depression/PTSD and their views on parenting. Parenting questions were chosen as it is found to be highly predictive of depression as seen in [13]. Moreover, only a few topics are discussed in most interviews, e.g., only 14 topics cover over 80% of the interviews. Hence, manual topic modelling is a better approach than automated methods such as LDA.

By following this simple approach, we were able to extract all utterances in a participant's transcript that corresponded to each of the 7 topics.

### 4.2. Topic-modelling based Data Augmentation

After the topics were labeled for each participant's text and audio data, the following steps are done for data augmentation (note that *m* denotes the minimum number of unique topics in a transcript and *n* denotes the minimum number of augmented participants to be created*)*:

1. Find the number of unique topics for each participant's transcript. If the number of topics found is greater than *m*, the participant will be selected for data augmentation.
2. Randomly select *n* combinations of at least *m* or more unique topics from the transcript. (The size of each combination must be less than the number of unique topics each participant has).
3. For each combination, obtain text and/or audio segments associated with each of the topics. These segments are randomly shuffled.
4. Concatenate the shuffled segments together to create a new audio sample/transcript for each of the *n* combinations.

Figure 2: The topic-modelling based data augmentation selects a participant's relevant topics (first column), associated transcripts (second column) and spectrogram (third column). Out of all the topics, a subset of the topics is taken as a new sample/transcript.

Data augmentation is only performed on the training set. The augmented set created using the original 107 participants (31 depressed, 76 non-depressed), consists of 534 audio samples and text transcript pairs (262 depressed, 272 non-depressed). This set is used to train our best performing audio-only and text-only models respectively. Our multi-modal model is trained on a new set of 307 audio samples and text transcript pairs (136 depressed, 171 non-depressed). This resulted in a more balanced distribution of depressed and non-depressed participants for all three modalities.

### 4.3. Fine-tuned Transformer

Transformer model is a relatively new network architecture solely based on attention mechanism, dispensing with recurrence and convolutions entirely [11]. The model is first trained on large amounts of data to perform language modelling in an unsupervised manner [17].

We followed a standard transfer learning paradigm for binary classification by fine-tuning the model's weights using our participants' transcripts to predict if the participant is depressed or not.

Our Transformer is made of multi-headed self-attention operation, followed by position-wise feedforward layers. The pre-trained model uses 6 layers consisting of 8 heads for the multi-headed attention with a hidden state size of 512 units each. The output is a binary classification: depressed or non-depressed.

In utilizing the model, we first tokenize each sentence of the participant as the following: start token, context token and lastly extract token to inform the end of the sentence. We input these strings of tokenized sentences into the Transformer.

### 4.4. 1-Dimensional Convolutional Neural Network

Convolutional Neural Networks (CNNs) are used to solve complex image-driven pattern recognition tasks [14]. A 1D CNN layer is a convolutional layer that convolves over only one dimension of the image, instead of the typical two dimensions.

For modelling audio, we used 1D CNNs to perform convolutions over the time dimension of mel-spectrograms for the participants. A mel-spectrogram is a 2D representation of time over frequency and can be visualized as an image. Temporal convolutions were chosen as two axes of a spectrogram have different meanings and units (time vs. frequency), unlike images (width and height in pixels) [15].

Another reason is that square convolution and pooling in 2D CNNs for image tasks can cause confusion among different audio classes and weaken the discriminative ability of any model applied to audio data [5]. 1D CNNs are very effective and derive interesting features from shorter (fixed-length) segments of the overall dataset, where the location of the feature within the segment is not of high relevance [15, 16].

Our model is made of 1D convolution layers, max pooling operations, and a global temporal pooling layer. Lastly, a multi-layer perceptron is used to predict the binary values: depressed or not depressed.

### 4.5. Multi-modal Fusion using Feedforward Network

We performed multi-modal fusion by training a simple feedforward model with both features: audio and text. We took the pre-trained text/audio models and used the values obtained at the last layer before the binary prediction layer as features which we concatenated together before feeding it to the model.

Following [6], each set of features were optimized with training data from their respective modality. Hence, the Transformer and the CNN are not restrained during fusion. Only the feedforward network's weights were learned.

### 4.6. Multi-modal Depression Classification

As there are S modalities in our collected samples, denoting $D^S \in R^{N \times D}$ and the corresponding features of the $s^{th}$ modality and the latent feature $B_n = [\alpha_n^1, \alpha_n^2, ..., \alpha_n^s] \in R^{D \times S}$ can be taken as the sparse representation of the $n^{th}$ sample $q_n$. We can train a binary classifier to detect depressed participants based on the following loss function:

$$\sum_{s=1}^{S} \sum_{n=1}^{NL} l(yn, Ws, \alpha) + \frac{p}{2} \sum_{s=1}^{S} ||W||$$

where W is the coefficient matrix, p is a regularization parameter, and $l_{su}(y_u, w^s, \alpha_n^s)$ measures how the classifier, parametrized by $w^s$, can predict $y_n$ by observing $\alpha_n^s$. Thus an optimization of the loss function can be realized by gradient methods.

## 5. EXPERIMENTS

### 5.1. Experimental Approach

We conducted a set of experiments on the text and audio features obtained from each participant to detect if they are depressed or not (binary classification). Compared to the current state-of-the-art [6, 18], we reported the results of the development set for the binary classification task using F1 score, precision, and recall. The experimental approach is as follows: we first evaluate the impact of data augmentation on the performance of deep learning models for audio and text modalities separately. Next, we evaluated the various models in a multi-modal setting with both text and audio features.

### 5.2. Text-only Model: Transformer

We experimented with three different approaches to train the Transformer model:

**1. Full Transcript Transformer (Trf-Full):** For each participant in both the training and testing set, we provided the model with a concatenated sequence containing all utterances by the participant in their transcript.

**2. Topic-modelling Based Transformer (Trf-Topic):** For each participant in the training and testing set, we provided a concatenated string containing only those utterances belonging to the 7 important topics found through topic modelling. We pad each utterance with a special topic token before concatenation.

**3. Data Augmentation Based Transformer (Trf-Augm):** For each participant in the training set, we followed the same methodology as in Trf-Topic and additionally performed topic-modelling based data augmentation to increase the amount of data and balance the two classes. Data augmentation is not performed on the test set.

With the same pre-trained model set up, all three models are fine-tuned with Adam optimizer (with learning rate of 6.25e-6) for 1 epoch each.

### 5.3. Audio-only Model: 1D CNN

We repeated our experiment with three different approaches to training the 1D CNN model:

**1. Full Audio CNN (CNN-Full):** For each participant in both the training and testing set, we provided the model with the Mel-spectrogram of each interview.

**2. Topic-modelling Based CNN (CNN-Topic):** For each participant in the training and testing set, we provided the model with the Mel-spectrogram that belongs to the 7 topics.

**3. Data Augmentation Based CNN (CNN -Augm):** For each participant in the training set, we followed the same methodology as in CNN-Topic and additionally perform topic-modelling based data augmentation to increase the amount of data and balance the two classes. For testing set, data augmentation was not performed.

All three models consisted of 4 1D convolution layers consisting of 64 filter of width 75 and stride 1. We used L2 regularization lambda of 0.01 and batch normalization after every convolution layer to reduce overfitting. Rectified linear units (ReLU) activation functions are used with a dropout rate of 0.5 in each layer. A max pooling operation with a pooling size of 2 is added between the layers. Max pooling is used to downsize the intermediate inputs. After the last convolution layer, a global temporal pooling layer was added. This layer is used to calculate the statistics of the learned features over the time axis using the concatenation of 3 different pooling strategies: L2-norm, mean and maximum. Lastly, a fully connected layer with 64 neurons and ReLU activation functions, and an output layer with a Softmax activation function is used. Each model is trained using the Adam optimizer (with learning rate of 6.25e-4) with a batch size of 30 for 1 epoch.

## 5.4. Multi-modal Model: Feedforward Network

We repeated our experiments with three different approaches to training the multi-modal model. Each feedforward model is provided with a concatenation of 512-dimensional text features and 64-dimensional audio features:

**1. Full Transcript and Audio model (Trf+CNN-Full):** For each participant in the training and testing sets, we concatenated features from Trf-Full for text and CNN-Full for audio.

**2. Topic-modelling Based Transcript and Audio model (Trf+CNN -Topic):** For each participant in the training and testing set, we concatenated features from Trf-Topic for text and CNN-Topic for audio.

**3. Data Augmentation Based Transcript and Audio model (Trf+CNN -Augm):** For each participant in the training and testing set, we concatenated features from Trf-Augm for text and CNN-Augm for audio.

In search of the optimal model, we explored the following hyperparameter spaces: number of hidden layer {0, 1, 2, 3}, dropout rate {0, 0.1, 0.5}, number of hidden nodes {32, 64, 128}, activation function {'ReLU', 'linear'}, Adam optimizer with learning rate {6.25e-3, 6.25e-4, 6.25e-5, 6.25e-6}, and number of epochs {1, 5, 10}. The best performing model comprised of 2 hidden layers for Trf+CNN-Full, 1 layer for Trf+CNN-Full and 3 layers for Trf+CNN-Aug, 128 hidden units in each layer (64 for Trf+CNN-Augm), linear activation function for all, 0.1 dropout rate and 6.25e-5 learning rate for all, and trained for 10 epochs (5 epochs for Trf+CNN-Full).

## 6. RESULTS AND DISCUSSION

The compiled results of all our experiments, as well as the published state-of-the-art models, are displayed in Table 1.[1]

---

[1] Code available at:
github.com/genandlam/multi-modal-depression-detection

---

For all three settings, models trained using our data augmentation framework outperform the previously published results. For the audio-only modality, our 1D CNN achieved an F1 score of 0.67 vs. 0.63 in [6]; for the text-only modality, our Transformer model achieved an F1 scores of 0.78 vs 0.76 in [18]; and for the multi-modal setting, our feedforward network achieved an F1 Score of 0.87 vs 0.77 in [6].

We believe that the performance gains achieved are due to the models being trained with more data that balanced the uneven class distribution. Secondly, training the models with consistent augmented datasets helped them pick up prominent features in both depressed and non-depressed participants, leading to a better performance.

We would like to note that the results in Williamson et al., 2016 (marked with * in Table 1) applied probability scaling to their model predictions [18]. Hence, we cannot directly compare our results with theirs, following [6].

Table 1: **Results** for state-of-the-art models and our approaches. Best in **bold**. (*) denotes results that cannot be directly compared.

| Model | Features Type | F1 | Precision | Recall |
|---|---|---|---|---|
| Williamson et al., 2016 | Audio | 0.50 | - | - |
| Williamson et al., 2016 | Text | 0.76 | - | - |
| Williamson et al., 2016 * | Text | 0.84 | - | - |
| Alhanai et al., 2018 | Audio | 0.63 | 0.71 | 0.56 |
| Alhanai et al., 2018 | Text | 0.67 | 0.57 | 0.80 |
| Alhanai et al., 2018 | Text + Audio | 0.77 | 0.71 | 0.83 |
| CNN-Full | Audio | 0.56 | 0.44 | **0.78** |
| CNN-Topic | Audio | 0.63 | 0.60 | 0.67 |
| CNN-Augm | Audio | **0.67** | **0.78** | 0.58 |
| Trf-Full | Text | 0.45 | 0.37 | 0.58 |
| Trf-Topic | Text | 0.71 | 0.55 | **1.0** |
| Trf-Augm | Text | **0.78** | **0.82** | 0.75 |
| Trf+CNN-Full | Text + Audio | 0.67 | 0.60 | 0.75 |
| Trf+CNN-Topic | Text + Audio | 0.69 | 0.64 | 0.75 |
| Trf+CNN-Augm | Text + Audio | **0.87** | **0.91** | **0.83** |

## 7. CONCLUSION

We proposed a novel method that incorporates context-aware and data-driven approaches for depression detection. We introduced a data augmentation procedure based on topic modelling and demonstrated its effectiveness for training deep learning models for audio and text modalities.

Our deep 1D CNN and Transformer models achieved state-of-the-art performance for audio and text modalities respectively (F1 score 0.67 vs 0.63, and 0.78 vs 0.76). Combining them in a multi-modal framework also outperforms state-of-the-art for the combined setting (F1 score 0.87 vs 0.77). Our strong experimental performance points to the promise of combining context-aware and data-driven methods for the automatic detection of depression.

Future work will focus on better multi-modal fusion for features from audio and text modalities, and on jointly training deep neural networks for multiple modalities.